\documentclass[journal]{IEEEtran}

\usepackage{cite}

\ifCLASSINFOpdf
  \usepackage[pdftex]{graphicx}
\else
  \usepackage[dvips]{graphicx}
\fi

\usepackage[cmex10]{amsmath}
\usepackage{subfigure}
\usepackage{amssymb}
\usepackage{multirow}
\usepackage{makecell}
\usepackage{marvosym}
\usepackage{url}
\usepackage{algorithm}
\usepackage{algorithmic}
\graphicspath{{Figure/}}

\begin{document}

\title{Temporally Identity-Aware SSD with Attentional~LSTM}

\author{Xingyu Chen, Junzhi Yu, \emph{Senior Member, IEEE,} and Zhengxing Wu

\thanks{
Manuscript received May 8, 2018; revised August 27, 2018 and November 23, 2018; accepted January 16, 2019. This work was supported in part by the National Natural Science Foundation of China under Grant 61725305, Grant 61633004, Grant 61633020, Grant 61633017, and Grant 61603388; and in part by the Beijing Natural Science Foundation under Grant 4161002. \emph{(Corresponding author: Junzhi Yu.)}	

X. Chen and Z. Wu are with the State Key Laboratory of Management and Control for Complex Systems, Institute of Automation, Chinese Academy of Sciences, Beijing 100190, China and University of Chinese Academy of Sciences, Beijing 100049, China (e-mail: chenxingyu2015@ia.ac.cn; zhengxing.wu@ia.ac.cn).

J. Yu is with the State Key Laboratory of Management and Control for Complex Systems, Institute of Automation, Chinese Academy of Sciences,
Beijing 100190, China, and also with the Beijing Innovation Center for Engineering Science and Advanced Technology, Peking University,
Beijing 100871, China (e-mail: junzhi.yu@ia.ac.cn).
}
}

\maketitle

\begin{abstract}
Temporal object detection has attracted significant attention, but most popular detection methods cannot leverage rich temporal information in videos. Very recently, many algorithms have been developed for video detection task, yet very few approaches can achieve \emph{real-time online} object detection in videos. In this paper, based on attention mechanism and convolutional long short-term memory (ConvLSTM), we propose a temporal single-shot detector (TSSD) for real-world detection. Distinct from previous methods, we take aim at temporally integrating pyramidal feature hierarchy using ConvLSTM, and design a novel structure including a low-level temporal unit as well as a high-level one (LH-TU) for multi-scale feature maps. Moreover, we develop a creative temporal analysis unit, namely, attentional ConvLSTM (AC-LSTM), in which a temporal attention mechanism is specially tailored for background suppression and scale suppression while a ConvLSTM integrates attention-aware features across time. An association loss and a multi-step training are designed for temporal coherence. Besides, an online tubelet analysis (OTA) is exploited for identification. Our framework is evaluated on ImageNet VID dataset and 2DMOT15 dataset. Extensive comparisons on the detection and tracking capability validate the superiority of the proposed approach. Consequently, the developed TSSD-OTA achieves a fast speed and an overall competitive performance in terms of detection and tracking. Finally, a real-world maneuver is conducted for underwater object grasping. The source code is publicly available at \url{https://github.com/SeanChenxy/TSSD-OTA}.
\end{abstract}


\IEEEpeerreviewmaketitle

\section{Introduction}
\IEEEPARstart{T}{aking} the advantage of convolutional neural network (CNN), existing detection methods commonly fall into two categories, i.e., one-stage and two-stage detectors. The former is represented by RCNN family \cite{bib:RCNN,bib:FastRCNN,bib:FasterRCNN,bib:MaskRCNN} and RFCN \cite{bib:RFCN}, all of which detect objects based on region proposal. On the other hand, YOLO \cite{bib:YOLO}, SSD \cite{bib:SSD}, RetinaNet \cite{bib:RetinaNet}, and etc., detect objects in a one-step fashion with a single-shot network. In particular, making use of CNN's features more effectively, SSD is one of the first methods that adopt the pyramidal feature hierarchy for detection. However, most works have largely focused on detecting in static images, ignoring temporal consistency in videos. Thus, it is imperative to develop an approach to integrate spatial features with temporal information. In addition, because of its relatively fast detection speed, the one-stage detector is more suitable for real-world applications.

Recurrent neural network (RNN) has seen great success in some sequence processing tasks \cite{bib:Zh17,bib:Su14}. Typically, long short-term memory (LSTM) is proposed for longer sequence learning \cite{bib:LSTM}. For spatiotemporal visual features, Shi \emph{et al.} developed a convolutional LSTM (ConvLSTM) to associate LSTM with spatial structure \cite{bib:ConvLSTM}. However, as for detection, selecting discriminative features for ConvLSTM is a pivotal step, because only a small part of visual features can devote themselves to detecting. Fortunately, attention is an exciting idea which imitates human's cognitive patterns, promoting CNN to concern on essential information. For example, Mnih~\emph{et al.} proposed a recurrent attention model to find the most suitable local feature for image classification \cite{bib:RAM}. Yet, very few literature has reported an attention-based temporal module for image-sequence detection.

As a closely relevant field, object tracking requires the initial position to be known as a prior knowledge \cite{bib:CorFilter}. Moreover, detection and tracking are moving towards unity in recent years, where a detector and a tracker tend to be cascaded in most up-to-date approaches \cite{bib:DT,bib:CDT}. This solution generally raises model complexity and computational cost. Hence, it is worthwhile to exploit a tracker-like detector for their in-depth combination and time efficiency. Tracking-by-detection is a popular idea, advocating that a detector should output tracker-like results, where the tracking component is actually designed for data association \cite{bib:ROLO,bib:ALSTM,bib:MDP,bib:SORT}. Nevertheless, the detector and tracking component are usually independent in existing tracking-by-detection framework, so it is essential to jointly investigate their performance.

\begin{figure}[!t]
\centering
\includegraphics[width=7cm]{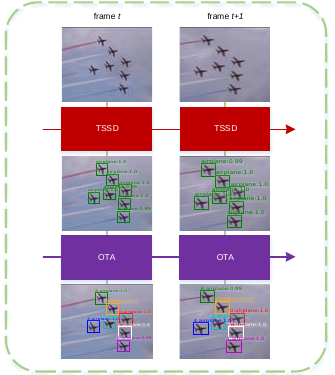}  
\caption{The main idea of TSSD-OTA. We aim to temporally detect objects in videos, and generate tracking-like results with low computational costs. The TSSD is a temporal detector, and the OTA is designed for identification.}
\label{fig:intro}
\end{figure}

As shown in Fig.~\ref{fig:intro}, our approach is able to detect objects across time and link them frame by frame using individual identities, and we call this kind of detection approach identity-aware detector. Aiming at detecting objects in videos, we propose a temporal detection model based on SSD, namely, temporal single-shot detector (TSSD). To integrate features across time, ConvLSTMs are employed for temporal information. Due to the pyramidal feature hierarchy for multi-scale detection, SSD always generates a large body of visual features with multi-scale semantic information, thus we design a new structure including a low-level temporal unit as well as a high-level one (LH-TU) for their temporal propagation. Furthermore, as for multi-scale feature maps, only a small part of visual features are related to objects. Thereby, attention mechanism is adopted for background suppression and scale suppression, then we propose an attentional ConvLSTM (AC-LSTM) module. Subsequently, an association objective and a multi-step method are developed for sequence training. Ultimately, an online tubelet analysis (OTA) is carried out for identification. As a consequence, the TSSD-OTA achieves considerably improved detection and tracking performance for consecutive vision in terms of both accuracy and speed. To the best of our knowledge, only few temporal one-stage detectors have been reported. Moreover, a real-time, online, and identity-aware detector is absent in existing detection frameworks. The contributions made in this paper are summarized as follows:
\begin{itemize}
  \item We design an LH-TU structure to effectively propagate pyramidal feature hierarchy across time. Moreover, we propose an AC-LSTM module as a temporal analysis unit, in which a temporal attention mechanism is designed for background suppression and scale suppression. Then, the corresponding training methods are developed for the TSSD.
  \item For identification, an OTA algorithm is exploited with the low-level AC-LSTM, which links detected objects frame by frame in a fairly fast speed.
  \item We achieve a considerably improved results on ImageNet VID dataset and 2DMOT15 dataset in terms of detection and tracking.
\end{itemize}

The remainder of this paper is organized as follows. Section~\ref{sec:RW} presents the related works. Thereafter, our approach including TSSD and OTA is elaborated in Section~\ref{sec:ME}. Next, Section~\ref{sec:EXP} provides the experimental results and discussion. Finally, conclusion and future work are summarized in Section~\ref{sec:CON}.

\begin{figure*}[!t]
\centering
\includegraphics[width=13cm]{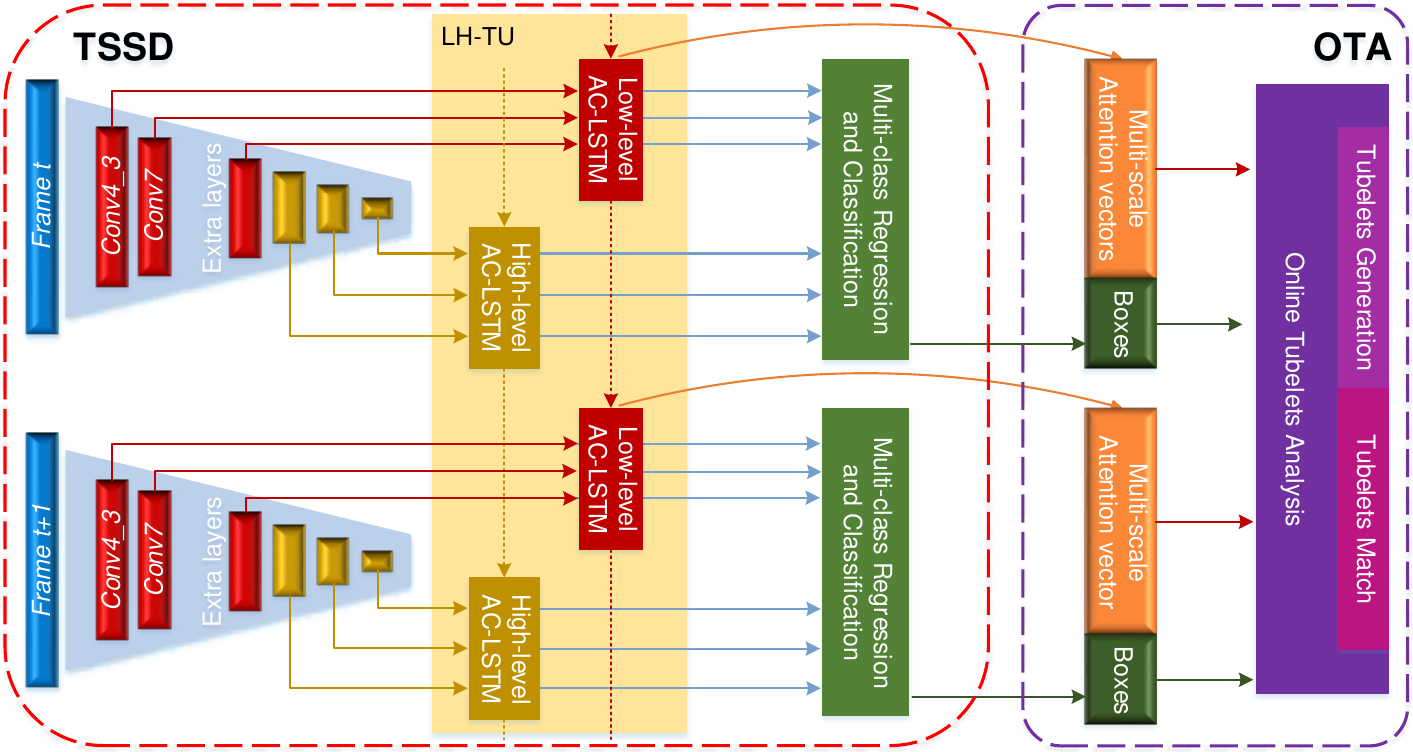}    
\caption{The schematic illustration of TSSD-OTA. The low-level features share an AC-LSTM and high-level features do so, namely, LH-TU. Next, the hidden states of ConvLSTM will be used for multi-box regression and classification. Eventually, based on multi-scale attention maps, online tubelet analysis is conducted for identification. Although we show a two-frame detection process here, the TSSD-OTA learns from all previous frames, and generates current hidden states using current pyramidal features as well as all previous memories. Moreover, the valid memory length is under the control of AC-LSTM's forget gate.}
\label{fig:TSSD}
\end{figure*}

\section{Related Work}
\label{sec:RW}
\subsection{Advancement of Detection Framework for Videos}
At the beginning, static detection and post-proposing methods are combined to counteract video detection task \cite{bib:VideoTub,bib:TCNN,bib:SeqNms}. They statically detect in each video frame, then comprehensively deal with multi-frame results. Kang \emph{et al.} developed detection methods based on tubelet, which is defined as temporally propagative bounding boxes in video snippet \cite{bib:VideoTub,bib:TCNN}. Their method TCNN contains still-image object detection, multi-context suppression, motion guided propagation, and temporal tubelet re-scoring. Taking inspiration from the non-maximum suppression (NMS), Han \emph{et al.} proposed an SeqNMS to suppress temporally discontinuous bounding boxes \cite{bib:SeqNms}. However, these solutions come with two major drawbacks. The complex post-processing could affect time efficiency, and they do not improve the performance of the detector itself.

Faster RCNN uses region proposal network for object localization \cite{bib:FasterRCNN}, so some approaches for video detection try to enhance the effectiveness of RPN with temporal information \cite{bib:CloseLoop,bib:VOP,bib:TPN,bib:STMN,bib:OL}. Galteri \emph{et al.} designed a closed-loop RPN to merge new proposals with previous detection results. This method effectively reduces the number of invalid regions, but it may also make the proposed regions excessively concentrated. Kang \emph{et al.} developed tubelet proposal networks (TPN) to generate tubelets rather than bounding boxes. Then, an encoder-decoder LSTM is used for classification. The TPN integrates temporal information, but it requires the future messages. Such methods are extended from two-stage detectors, resulting in relatively low computational efficiency.

\subsection{Detection and Tracking}
Feichtenhofer \emph{et al.} associated an RFCN detector with a correlation-filter-based tracker \cite{bib:CorFilter} to detect objects in videos, called D\&T \cite{bib:DT}. Thanks to the tracking method, it achieves a high recall rate, but this cascaded system could seriously increase the model complexity, and impair the inference speed.

In terms of tracking-by-detection, Xiang \emph{et al.} converted the tracking task to decision making, and their policy relies on tracking states \cite{bib:MDP}. H.~Kim and C.~Kim combined a detector, a forward tracker, and a backward tracker for tracing multiple objects in video sequences, and the detector was also used to refine tracking results \cite{bib:CDT}. Ning \emph{et al.} proposed an ROLO based on YOLO and LSTM for tracking \cite{bib:ROLO}. The YOLO is responsible for static detection, and the visual features as well as positions of high-score objects will be fed to LSTM for temporally modeling. Lu \emph{et al.} proposed association LSTM (ALSTM) to temporally analyze relations of high-score objects, and an association loss was designed for identification \cite{bib:ALSTM}. Pragmatically, Bewley \emph{et al.} developed an SORT with Kalman filter \cite{bib:Kalman} and Hungarian method \cite{bib:Hungarian} for real-time tracking \cite{bib:SORT}, and the tracking component achieves a speed of $260$ frames per second (FPS). Nevertheless, in most of these methods, the visual features in detector have not been efficiently used yet.

\subsection{RNN-Based Detector}
Very recently, up-to-date approaches have associated the detector with RNN. Liu and Zhu reported a mobile video detection method based on SSD and LSTM, called LSTM-SSD \cite{bib:LSTM_SSD}. Moreover, they also designed a Bottleneck-LSTM structure to reduce computational costs. As a result, LSTM-SSD reached a real-time inference speed of up to $15$ FPS on a mobile CPU. Xiao and Lee developed a spatial-temporal memory module (STMM) with ConvGRU  \cite{bib:ConvGRU} for temporal information propagation  \cite{bib:STMN}. In particular, ``MatchTrans" was proposed to suppress the redundant memory.

\section{Approach}
\label{sec:ME}
In this section, we firstly present the proposed architecture, including the LH-TU and the AC-LSTM. Then, we describe how to train the network in detail. Finally, the methodologies of OTA algorithm will be briefed.

\subsection{Architecture}
Extending form SSD with VGG-16 \cite{bib:VGG} as the backbone, we build a temporal architecture, where \emph{fc6, fc7} in original VGG-16 are converted to convolutional layers, namely, \emph{Conv6, Conv7}. Referring to Fig.~\ref{fig:TSSD}, the proposed TSSD is based on forward CNN and RNN that generate pyramidal features for detection. Convolutional detection head is designed for multi-class classification and regression (shown in green). In this process, convolution operations are leveraged to predict objects¡¯ information with multi-scale visual features, then a fixed number of bounding boxes and the category-discriminative scores indicating the presence different classes of objects on those boxes, followed the NMS to generate the final results. The spatial resolution of the input image is $300\times 300$. \emph{Conv4\_3, Conv7, Conv8\_2, Conv9\_2, Conv10\_2, Conv11\_2} are employed as pyramidal features, whose size are $38\times 38\times 512$, $19\times 19\times 512$, $10\times 10\times 512$, $5\times 5\times 256$, $3\times 3\times 256$, and $1\times 1\times 256$, respectively. As for sequence learning, the TSSD is equipped with multi-scale feature-integration structures, i.e. the LH-TU and the AC-LSTM. The LH-TU takes aim at propagation of pyramidal feature hierarchy, whereas the AC-LSTM aims to effectively produce temporal memory without useless information.

\subsubsection{LH-TU}
We use the same two structures to integrate the pyramidal feature hierarchy temporally, called LH-TU. There are pyramidal features for six-scale semantic information in adopted SSD model, and their feature sizes are diverse from each other. In original SSD framework, there are $512,1024,512,256,256,256$ channels in feature maps from low-level to high-level. Creatively, we divide the multi-scale feature maps into two categories according to their  hierarchical relation and channel sizes, i.e., low-level features and high-level features. That is, on one hand, we group the multi-scale feature maps according to the order of different convolutional layers. It is known that convolution operation extracts visual features gradually, and shadow features contain more image details while the high-level features cover more semantic information. Thus, the LH-TU is in favor of the learning process. On the other hand, the low-level and high-level features should share their respective temporal unit, but a ConvLSTM unit cannot process channel-size-variable features, so in the manner of LH-TU, we only change \emph{Conv7}'s channel size to $512$ to remain original SSD structure as much as possible. Therefore, as illustrated in Fig.~\ref{fig:TSSD}, we treat the first three feature maps as low-level features (shown in red), whereas the last three maps are considered as high-level features (shown in gold). The channel sizes of low-level features are unified as $512$ to share a temporal unit while that of high-level features remain as $256$. Additionally, the low-level features cover more details, whereas the high-level features contain more semantic information. Correspondingly, the LH-TU including a low-level temporal unit and a high-level one is designed for them.

\begin{figure}[!t]
\centering
\includegraphics[width=7cm]{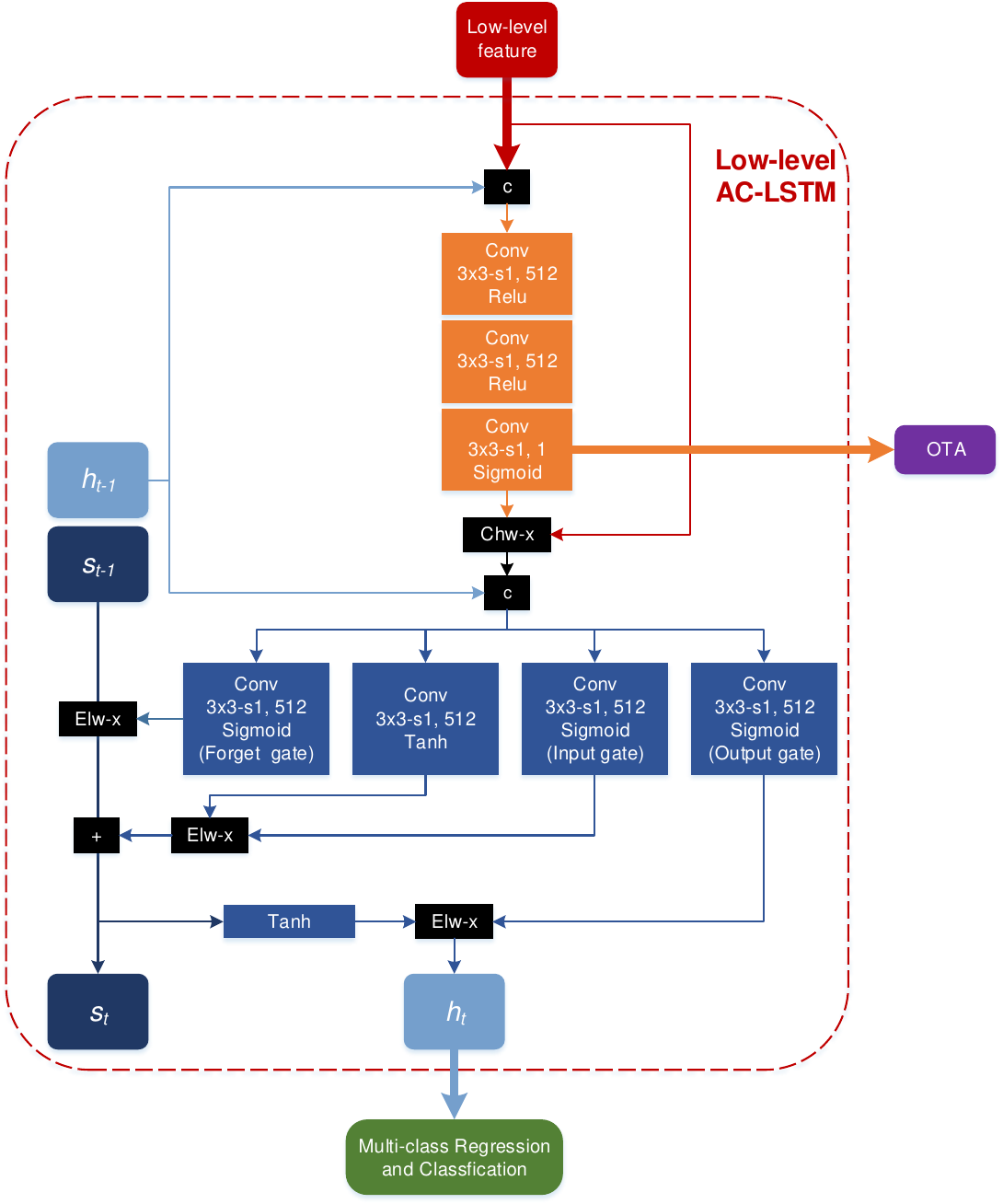}   
\caption{Implementation detail of AC-LSTM. ``c'' denotes concatenation; ``Chw-x'', ``Elw-x'' represent channel-wise and element-wise multiplication; ``+'' is element-wise summation. The generated hidden state is used for detection while the attention map is employed by both AC-LSTM and OTA.}
\label{fig:detail}
\end{figure}

\subsubsection{AC-LSTM}
In object detection task, most features are related to background. Moreover, feature maps in different scales contribute differently to the detection. Therefore, it is inefficient when a ConvLSTM handles background or aforementioned small-contributed multi-scale feature maps. For example, if an object's size is too small, its detection will be contributed by \emph{Conv4\_3}, in which features associated with the small object are far less than that for background. Moreover, all the higher-level feature maps can be considered useless, which should be suppressed to avoid the false positive. To that end, we propose an AC-LSTM for background suppression and scale suppression, in which a temporal attention mechanism selects object-aware features for a ConvLSTM, and in turn, the ConvLSTM provides the attention mechanism with temporal information to improve attention accuracy. As a temporal analysis unit, the AC-LSTM can be formulated as:
\begin{equation} \label{eqn:ConvLSTM}
\begin{array}{l}
a_t=\text{sigmoid}(W_a*[x,h_{t-1}]) \\
i_t=\text{sigmoid}(W_i*[a_t\circ x,h_{t-1}]+b_i) \\
f_t=\text{sigmoid}(W_f*[a_t\circ x,h_{t-1}]+b_f) \\
o_t=\text{sigmoid}(W_o*[a_t\circ x,h_{t-1}]+b_o) \\
c_t=\text{sigmoid}(W_c*[a_t\circ x,h_{t-1}]+b_c) \\
s_t=(f_t\odot s_{t-1})+(i_t\odot c_t) \\
h_t=o_t\odot\tanh(s_t),
\end{array}
\end{equation}
where $*$ denotes convolution; $[\cdot,\cdot]$ is concatenation; $\odot$ is element-wise multiplication; and $\circ$ represents that a one-channel map multiplies with each channel in a multi-channel feature map. At time step $t$, $a_t, h_t, i_t, f_t, o_t, c_t, s_t$ are attention map, hidden state, input gate, forget gate, output gate, LSTM's incoming information and memory, respectively.

As shown in Fig.~\ref{fig:detail}, the AC-LSTM is designed with CNN and RNN. Current feature map ($x$) and previous hidden state ($h_{t-1}$) serve as the input of the attention module. After a three-layer convolution, a one-channel attention map ($a$) is generated containing pixel-wise positions for object-aware features, which will be used to select useful features in the AC-LSTM and identify objects in the OTA. It should be mentioned that, instead of binary selection ($1$ for attention area and $0$ otherwise), each element in the attention map is continuous in $[0, 1]$, so as to describe object's saliency mass more effectively. For feature selection, each channel of current feature map multiplies this attention map pixel-by-pixel, and the attention-aware feature ($a\circ x$) can be obtained.  The attention-aware feature and previous hidden state are concatenated as the input of the ConvLSTM. Different from traditional LSTM, gates ($i$, $f$, $o$) and incoming information ($c$) will be computed with convolution operation \cite{bib:ConvLSTM}. Subsequently, controlled by gates, the temporal memory ($s$) will be updated, and current hidden state is generated for multi-box regression and classification. During this operation, $a\circ x,i,f,o,c,s,h$ are in the same size. In addition, we also use dropout regularization \cite{bib:DropOut} for the attention-aware feature during training. Apparently, temporal information transmission is conducted twice in temporal attention and ConvLSTM.

Note that the attention mechanism and input gate play different roles, although both of them can be aware of useful features. For background suppression and scale suppression, the attention mechanism works for spatial location in each 2D map, whereas the input gate can deal with the 3D feature along the channel to preserve discriminative data.

\subsection{Training}
We design a multi-task objective to train the TSSD, including a localization loss $\mathcal L_{loc}$, a confidence loss $\mathcal L_{conf}$, an attention loss $\mathcal L_{att}$, and an association loss $\mathcal L_{asso}$,
\begin{equation} \label{eqn:Obj}
\mathcal L = \frac{1}{M}(\alpha\mathcal L_{loc} + \beta\mathcal L_{conf}) + \gamma\mathcal L_{att} + \xi\mathcal L_{asso},
\end{equation}
where $M$ is the number of matched boxes, and $\alpha,\beta,\gamma,\xi$ are trade-off parameters. $\mathcal L_{loc}$ and $\mathcal L_{conf}$ are defined in accordance with SSD \cite{bib:SSD}. Then, we train the TSSD in three steps.

\subsubsection{Attention Loss}
The generation of attention maps is supervised using cross entropy. At first, we construct the ground truth attention map $A_g$, in which elements in ground truth boxes equal to $1$ and others are $0$. There are six feature maps for multi-box prediction, which generate multi-scale attention maps $A_{p_{sc}}$. Therefore, each $A_{p_{sc}}$ is firstly unified to the same resolution as the input image through bilinear upsampling operation, followed by the generation of $A^{up}_{p_{sc}}$. Each upsampled attention map can generate a scale-related attention loss with cross entropy, and we add up 6-scale losses as the final attention loss. Then, $\mathcal L_{att}$ can be given as
\begin{equation} \label{eqn:Att}
\mathcal L_{att} = \sum_{sc=1}^6 \mu(-A^{up}_{p_{sc}}\log(A_g)-(1-A^{up}_{p_{sc}})\log(1-A_g)),
\end{equation}
where $\mu$ averages all elements in a matrix.

\subsubsection{Association Loss}
Pixel-level changes could significantly impact the detection results, so an object in video always encounters large confidence fluctuations with a static detection method (studied in \cite{bib:TCNN}). Thus, towards temporal consistency of videos, an association loss should be developed for sequence training. To that end, we encourage the TSSD to generate similar global classification results for consecutive frames. We firstly compute top $k$ high confidence scores per class after NMS, then sum them up to generate a class-discriminative score list ($sl$). The score list should remain small fluctuation in consecutive frames. Thereby, the $\mathcal L_{asso}$ can be obtained,
\begin{equation} \label{eqn:Asso}
\mathcal L_{asso} = (\sum_{t=1}^{seq\_len}sl_t-sl_{ave})/seq\_len,
\end{equation}
where $sl_t$ is the score list at time step $t$; $sl_{ave}$ denotes the mean score list among $sl_{1:t-1}$; and $seq\_len$ represents the sequence length. It should be remarked that our proposed association loss works in a self-supervision manner. That is, there is no incoming ground truth label when computing $\mathcal L_{asso}$.

\subsubsection{Multi-Step Training}
We first train an SSD model following \cite{bib:SSD}. In the next step, the TSSD is trained based on well-trained SSD. We freeze the weights in the network except for AC-LSTM and detection head. In particular, the ConvLSTM is trained with RMSProp \cite{bib:RMSProp} while the rest of TSSD is trained using SGD optimizer with an initial learning rate of $10^{-4}$ and a decay rate of $0.1$ for $40$ epochs. The learning rate drops at the $30$th epoch. On the other hand, the TSSD should be trained with a sequence of frames, but the frame rates of videos are inconstant. Moreover, the motion speed of objects in videos varies considerably. For better generalization, it should not be trained frame by frame. Instead, we only choose $seq\_len$ frames in a video for backpropagation in an iteration. The $seq\_len$ frames are chosen uniformly based on the start frame $sf$ and skip $sp$, namely, random skip sampling (RSS), i.e.,
\begin{equation} \label{eqn:seq}
\begin{array}{l}
sp=R[1, v/seq\_len] \\[10pt]
sf=R[1, v-seq\_len\times sp+1],
\end{array}
\end{equation}
where $v$ is the total number of frames in a video, and $R[min, max]$ represents the operation of selecting an integer randomly between $min$ and $max$. Finally, the uniform $seq\_len$ frames are chosen with $sf$ as the start frame and $sp$ as the skip. In this paper, $seq\_len=8$. At this step, the association loss $\mathcal{L}_{asso}$ is not involved.

Thirdly, the full objective including $\mathcal L_{asso}$ is used to fine tune parameters for $10$ epoches. At this step, the learning rate is $10^{-5}$, and $sp=1$. The hyper parameters $\alpha=1,\beta=1,\gamma=0.5,\xi=2,\delta=3$ are selected based on model performance.

\subsection{Inference}
At inference phase, the LH-TU and the AC-LSTM integrate features across time, generating temporally-aware hidden state for regression and classification. Finally, we apply the NMS with jaccard overlap (IoU) of $0.45$ (for ImageNet VID) or $0.3$ (for 2DMOT15) per class and retain the top $200$ (for ImageNet VID) or $400$ (for 2DMOT15) high confident detections per image to produce the final detections.
\begin{algorithm}[!b]
\caption{Online Tubelet Analysis}\label{code:OTA}
\begin{algorithmic}[1]
\FOR{$cls \in$ classes}
\IF{$tubs[cls]$ is not empty}
\FOR{$obj \in$ detection result}
\STATE $S^{max}_{obj}=0$
\FOR{$tub \in tubs[cls]$}
\STATE $\mathcal S_{obj,tub}=$(\ref{eqn:simi})
\IF{$\mathcal S_{obj,tub} > S^{max}_{obj}$}
\STATE $S^{max}_{obj} = \mathcal S_{obj,tub}$
\STATE $candidate=tub$
\ENDIF
\ENDFOR
\IF{$S^{max}_{obj} > \mathcal T$}
\STATE $obj=obj^{candidate[id]}$
\ENDIF
\ENDFOR
\FOR{$tub \in tubs[cls]$}
\IF{len($obj^{tub[id]}$)$>$ 1}
\STATE preserve only one with maximal $\mathcal S$
\ENDIF
\ENDFOR
\ENDIF
\FOR{$obj \in \{obj^{-1}\}$}
\IF{$obj[conf]>\mathcal G$}
\STATE $obj=obj^{new\_id}$
\ENDIF
\ENDFOR
\STATE update $tubs[cls]$
\ENDFOR
\end{algorithmic}
\end{algorithm}

\subsection{Online Tubelet Analysis}
For online tracking-by-detection task, it is expected that the TSSD is endowed with the ability to identify objects. Tubelet in videos has been studied by \cite{bib:TCNN,bib:TPN,bib:OL}, each of which has a unique identity (ID). In the scope of detection, \cite{bib:TCNN} generates tubelets by tracking, whereas \cite{bib:TPN} and \cite{bib:OL} use tubelet proposal for batch-mode detection. However, these methods are either computationally expensive or not online. Thereby, we attempt to exploit a real-time online object association method based on tubelet for tracking-by-detection.

Features in a detector are not suitable for identification, because they always contain within-class-similar information. Fortunately, there are category-independent features in our framework, i.e., attention maps. Instead of being discriminative for each class, the attention maps describe the object's saliency mass at different visual scales. Thereby, the attention space is employed for fast data association. This idea is straightforward and computationally tractability, where the key intuition is that the saliency mass of objects are diverse from each other, and the attention mechanism is able to capture this subtle distinctiveness. The low-level AC-LSTM is employed for this task, because low-level features cover more detailed information. In short, attention maps have three merits for identification:
\begin{itemize}
  \item We no longer need extra instance-related training.
  \item They are instance-aware features that describe objects' saliency mass.
  \item Low computational costs are produced. That is, when the spatial size is sampled as $7\times 7$ for each object, the length of an attention vector is $147$, whereas it increases to $75267$ if low-level features or corresponding hidden states are employed without any other operation.
\end{itemize}

Hence, the attention similarity $as_{ij}$ between two objects $i$ and $j$ and the can be formulated as a cosine distance,
\begin{equation} \label{eqn:as}
as_{i,j} = \frac{av_i\cdot av_j}{||av_i||||av_j||},
\end{equation}
where $av$ denotes the $147$-dimension attention vector flattened by bilinear-sampled multi-scale attention maps.

Moreover, identification in a video can leverage more temporal coherence, so we also employ IoU $o$ in the OTA. Due to multiple objects in a tubelet, the similarity $\mathcal S_{obj,tub}$ between an object $obj$ and a tubelet $tub$ can be given as
\begin{equation} \label{eqn:simi}
\begin{array}{l}
as_{obj,tub} = (\sum_{k\in tub}as_{obj,k})/tub\_len \\[10pt]
o_{obj,tub} = \text{IoU}(obj,tub[0]) \\[10pt]
\mathcal S_{obj,tub} = \exp(o_{obj,tub})\times as_{obj,tub},
\end{array}
\end{equation}
where $tub\_len$ is current length of the tubelet, and $tub[0]$ denotes the most recent object.

Define $\mathcal G, \mathcal T$ to denote the tubelet generation score, match threshold, respectively. Suppose that each detected object is described as $obj[conf,loc,av]$, and each class-distinct tubelets set is denoted as $tubs[cls]$, each tubelet in which is given as $tub[id,objs]$, the OTA algorithm is presented in Algorithm~\ref{code:OTA}, where $obj^{id}$ denotes an object with an identify $id$, and $id=-1$ represents an object without an identify. $\text{len}(\cdot)$ computes the number of elements. Note that the maximal existence time after disappearance and maximal tubelet length are restricted when updating tubelets.

\section{Experiments and Discussion}
\label{sec:EXP}

\subsection{Dataset}
\subsubsection{ImageNet VID}
We evaluate the TSSD on ImageNet VID dataset \cite{bib:ImageNet}, which is the biggest dataset for temporal object detection now. The task requires algorithms to detect $30$-class targets in consecutive frames. There are $4000$ videos in the training set, containing $1181113$ frames. On the other hand, the validation set compasses $555$ videos, including $176126$ frames. We measure performance as mean average precision (mAP) over the $30$ classes on the validation set following \cite{bib:FasterRCNN,bib:SSD}. In addition, ImageNet DET dataset is employed as training assistance. The $30$ categories in VID dataset are a subset of the $200$ categories in the DET dataset. Therefore, following \cite{bib:VideoTub,bib:TCNN,bib:TPN,bib:DT}, we train the SSD with VID and DET (only using the data from the $30$ VID classes). In reality, there are millions of frames in VID training set, so it is hard to train a network directly using them. Additionally, the data for each category are imbalanced, because there are long videos (contain more than $1000$ frames) and short videos (contain only a dozen frames). Thus, following \cite{bib:DT}, we sample at most $2000$ images per class from DET, and select $10$ frames in each VID video for SSD training at the first step. In the second and third training steps, all the VID training videos are adopted.

\subsubsection{2DMOT15}
The TSSD-OTA is an identity-aware detector, so 2DMOT15 dataset \cite{bib:MOT} is employed to evaluate tracking performance. This is a  multi-object tracking consisting of $11$ training sequences. Since the annotations are available for the training set only, we split $5$ videos as the validation set following \cite{bib:ALSTM,bib:MDP}. In addition, another $3$ video sequences in MOT17Det \cite{bib:MOT16} dataset are employed for training.

\subsection{Runtime Performance}
Our methods are implemented under the PyTorch framework. The experiments are carried out on a workstation with an Intel 2.20 GHz Xeon(R) E5-2630 CPU, NVIDIA TITAN-X GPUs, 64 GB of RAM, CUDA 8.0, and cuDNN v6. The inference time is described in Table~\ref{tab:FPS}, and we achieve a beyond real-time speed for temporal detection or tracking.

\begin{table}[H]
\renewcommand{\arraystretch}{1.2}
\caption{FPS list on employed datasets by the proposed methods.}
\label{tab:FPS}
\centering
\begin{tabular}{c | c c }
\Xhline{1.5pt}
Method                & VID (FPS)& 2DMOT15 (FPS)  \\
\hline
SSD                   &$\sim45$ & $\sim77$  \\
TSSD    &$\sim27$ & $\sim30$ \\
TSSD-OTA    &$\sim21$ & $\sim27$ \\
\Xhline{1.5pt}
\end{tabular}
\end{table}

\subsection{Ablation Study on ImageNet VID}

\subsubsection{LH-TU}
Our proposed LH-TU is effective in the following aspects. Firstly, redundant parameters are avoided. For example, the original SSD contains $2.6$~M parameters, and SSD with LH-TU has $4.9$~M parameters. However, if six ConvLSTMs are employed for each feature map, the parameter size will dramatically increase to $15.5$~M, which leads to unstable training. Secondly, as reported in \cite{bib:ALSTM}, \emph{Conv4\_3} and \emph{Conv11\_2} make relatively less contribution to detection. That is, there are a small amount of data for oversized or tiny-size objects. Thus, the highest-level and lowest-level ConvLSTMs can hardly be well trained, if six-scale ConvLSTMs are employed. We find that the mAP increases by $0.92\%$ when LH-TU is adopted using two ConvLSTMs as temporal units due to the feature integration brought by ConvLSTMs.

\begin{figure}[!t]
\begin{center}
\includegraphics[width=7cm]{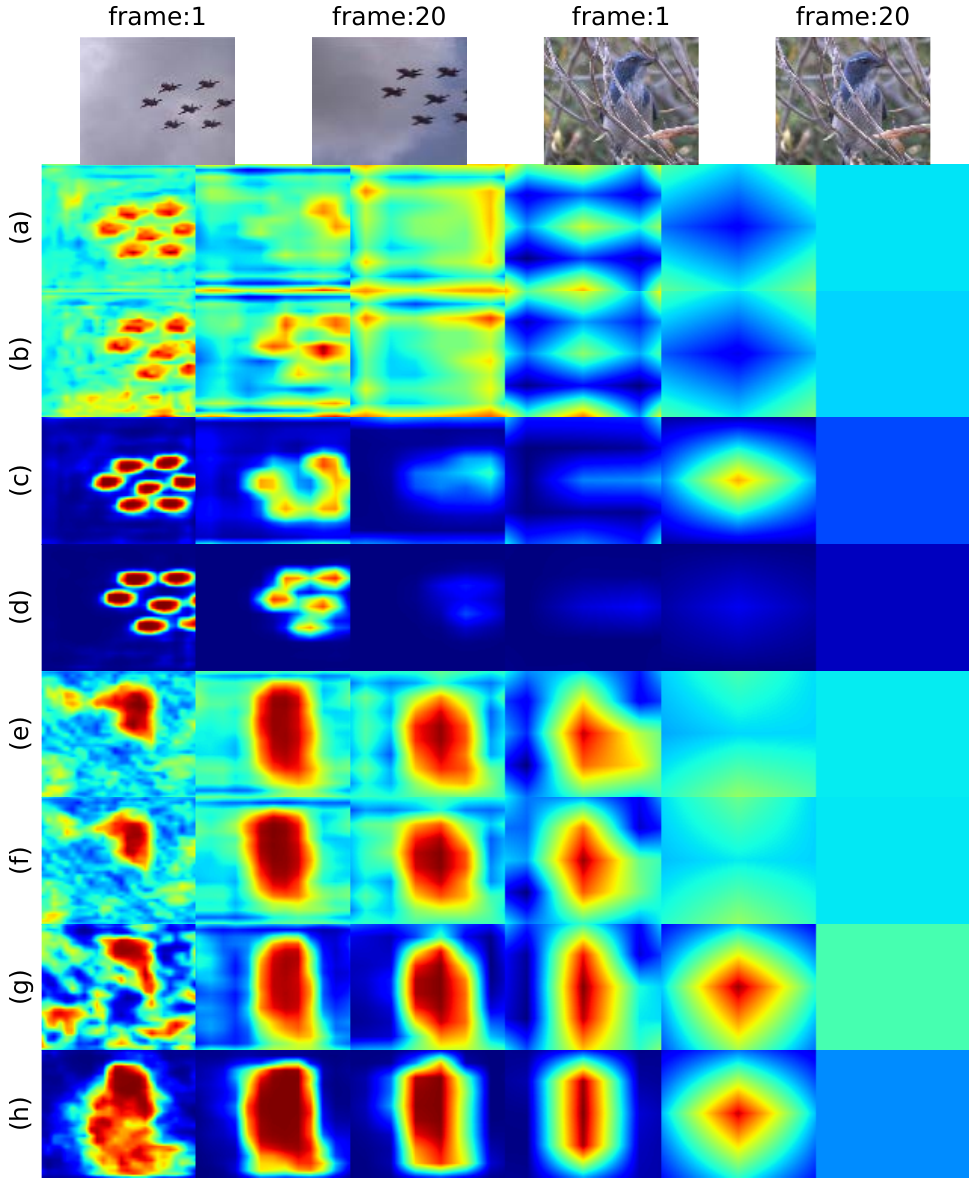}    
\caption{Effect of the ConvLSTM for attention mechanism. There are two video snippets containing small objects (airplane) or wild environment (bird). The traditional attention and temporal attention mechanism are used to generate multi-scale attention maps, in which crimson denotes higher level of concern while mazarine represents something neglected. (a)--(b) attention maps for airplanes generated by traditional module; (c)--(d) attention maps for airplanes generated by temporal module; (e)--(f) attention maps for bird generated by traditional module; (g)--(h) attention maps for bird generated by temporal attention method; In above $4$-pair maps, the former is for the first frame while the latter is with respect to the $20$th frames. Each line in (a)--(h) is multi-scale attention maps. From left to right, they are generated with \emph{Conv4\_3, Conv7, Conv8\_2, Conv9\_2, Conv10\_2, Conv11\_2}, respectively.}
\label{fig:att}
\end{center}
\end{figure}

\subsubsection{AC-LSTM}
\begin{figure}[!t]
\begin{center}
\includegraphics[width=7cm]{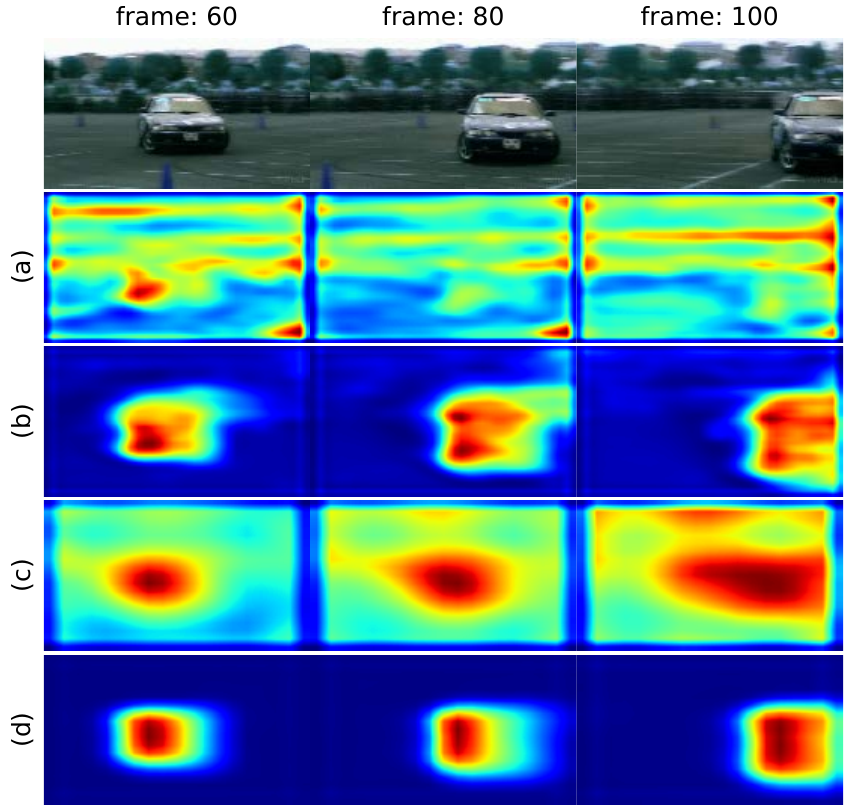}    
\caption{Effect of the temporal attention for ConvLSTM. The ConvLSTM's memory ($s$) is visualized (by computing the L2 norm across feature channels at each spatial location to get a saliency map). (a) original ConvLSTM's memory for \emph{Conv4\_3}; (b) ConvLSTM's memory for \emph{Conv4\_3} in AC-LSTM; (c) original ConvLSTM's memory for \emph{Conv7}; (d) ConvLSTM's memory for \emph{Conv7} in AC-LSTM;.}
\label{fig:cell}
\end{center}
\end{figure}

At first, we qualitatively analyze the interaction of the attention mechanism and ConvLSTM. As shown in Fig.~\ref{fig:att}, the comparison of temporal and traditional attention mechanism are presented. Note that the traditional attention only uses current feature map as the input. As for presented heat maps in Fig.~\ref{fig:att}, crimson means a higher probability of being a target, whereas the mazarine indicates background features. Moreover, multi-scale attention maps are generated in the TSSD, and the righter maps are generated with higher-level features. For the ease of observation, all maps in Figs.~\ref{fig:att} and \ref{fig:cell} have been unified to the same spatial resolution as the input image through a bilinear upsampling operation.

We choose two challenging scenes for this test. The airplane frames include small objects, and the bird frames contain rich stripes, both of which are difficult for attention operation. As delineated in Fig.~\ref{fig:att}(a), (b), (e), (f), the original attention method is not able to handle these two scenes. That is, although the targets are focused roughly, the background and small-contributed multi-scale feature maps are not suppressed effectively. Moreover, there is no improvement across time series. On the contrary, as illustrated in Fig.~\ref{fig:att}(c), (d), (g), (h), the proposed attention mechanism performs better. In short, our method not only localizes the targets more accurately, but also suppresses the background more efficiently. Further, our method is effective for scale suppression. For instance, small objects in airplane frames are described by \emph{Conv4\_3}. As delineated in Fig.~\ref{fig:att}(d), the attention maps for \emph{Conv4\_3,Conv7} localize airplanes, and all the last four maps are ``cold''. That is, when it comes to larger scale, our attention mechanism cannot find any target, so the whole of feature map has been suppressed. In addition, the performance of proposed approach improves along with the accumulation of temporal information. For example, in Fig.~\ref{fig:att}(g), (h), the attention map for \emph{Conv4\_3} can hardly find the bird in the first frame, but the bird's profile is focused without overmuch background in the $20$th frame. Moreover, if attention maps for the first frame are compared, a conclusion can be drawn that the temporal attention method is better even though the temporal information has not generated. The reason is that the temporal attention can be trained more effectively.

\begin{table}[!t]
\renewcommand{\arraystretch}{1.2}
\setlength\tabcolsep{4pt}
\caption{Effectiveness of various designs. All model are trained and validated on ImageNet VID dataset.}
\label{tab:mAP}
\centering
\begin{tabular}{c | c c c c c c}
\Xhline{1.5pt}
Component & \multicolumn{6}{c}{TSSD}  \\
\hline
Association loss?       & \checkmark &            &           &            &            &              \\
The 3rd training stage? & \checkmark &\checkmark  &           &            &            &               \\
RSS in the 2nd stage?   & \checkmark &\checkmark  &\checkmark &            &            &               \\
AC-LSTM?                & \checkmark &\checkmark  &\checkmark &\checkmark  &            &               \\
LH-TU?                  & \checkmark &\checkmark  &\checkmark &\checkmark  &\checkmark  &               \\
\hline
mAP(\%)           &  $65.43$    & $65.13$ & $64.76$     &$64.63$     &  $63.95$         & $63.03$  \\
\Xhline{1.5pt}
\end{tabular}
\end{table}

\begin{table*}[!t]
\renewcommand{\arraystretch}{1.2}
\setlength\tabcolsep{5pt}
\caption{Comparison of the TSSD and several prior and contemporary approaches.}
\label{tab:comp}
\centering
\begin{tabular}{c| c c c c c c | c c c }
\Xhline{1.5pt}
\multirow{2}{*}{Method} &          \multicolumn{6}{c|}{Components} &  \multicolumn{3}{c}{Performances} \\
                                 &Backbone    & one-stage?     & Optical flow?      & Tracking? & Attention? & RNN?    & Real time?    & ID?  & mAP  \\
\hline
\emph{offline methods} &&&&&& \\
STMN\cite{bib:STMN}              &VGG-16         &&              &          &           &\checkmark&           &             & 55.6 \\
TPN \cite{bib:TPN}               &GoogLeNet \cite{bib:GoogleNet}&&              &          &           &\checkmark&           &            & 68.4 \\
FGFA\cite{bib:FGFA}              &ResNet-101 \cite{bib:ResNet}  &&\checkmark    &          &           &          &           & & 76.3 \\
HPVD \cite{bib:HPVD}             &ResNet-101     &&\checkmark    &          &           &          &           &    & 78.6 \\
STSN \cite{bib:STSN}             &ResNet-101     &&              &          &           &          &           &    & 78.7 \\
Object-link\cite{bib:OL}         &ResNet-101     &&              &          &           &          &           &   & 80.6 \\
\hline
\hline
\emph{online methods} &&&&&&\\
Closed-loop \cite{bib:CloseLoop} &VGG-M \cite{bib:VGG_M}         &&          &           &          &           &           &   & 50.0 \\
Seq-NMS \cite{bib:SeqNms}        &VGG-16         &&              &          &           &          &           &   & 52.2 \\
LSTM-SSD\cite{bib:LSTM_SSD}      &MobileNet \cite{bib:MobileNet} &\checkmark &              &          &           &\checkmark&\checkmark & & 54.4 \\
TCNN \cite{bib:TCNN}             &DeepID+Craft \cite{bib:Deepid,bib:Carft} &&\checkmark    &\checkmark&           &          &           &  & 61.5 \\
D\&T \cite{bib:DT}               &ResNet-101 &     &          &\checkmark&           &          &           &   & 78.7 \\
TSSD(-OTA)                       &VGG-16     &\checkmark      &              &          &\checkmark &\checkmark&\checkmark  &\checkmark & 65.4 \\
\Xhline{1.5pt}
\end{tabular}
\end{table*}

On the other hand, there are benefits of the AC-LSTM over traditional ConvLSTM. Referring to Fig.~\ref{fig:cell}, the ConvLSTM's memories for \emph{Conv4\_3, Conv\_7} are visualized. Small-scale memory deals with image details, but the ConvLSTM cannot generate valid memory because it is confused by mussy features, whereas the AC-LSTM is able to memorize the essential information of the car without background (see Fig.~\ref{fig:cell}(a--b)). In consideration of the scale of the car, \emph{Conv7} is more crucial to detecting it. As delineated in Fig.~\ref{fig:cell}(c), the ConvLSTM for \emph{Conv7} has a tendency to learn trivial representations that just memorizes the inputs. Moreover, this memorization also involves the background. Thus, not all these information is useful for future detection, and they may incur inaccuracies. On the contrary, the AC-LSTM produces more clear memory with pivotal features (see Fig.~\ref{fig:cell}(d)). Nevertheless, we cannot draw a conclusion about the best temporal learning length. Thus, the AC-LSTM learns from all previous frames, and under the control of its forget gate, the length of the valid memory depends on sequential learning in a specific scenario. Owing to the above reasons, the improvement brought by AC-LSTM is evident, i.e., the mAP rises by $1.60\%$ based on SSD.

\subsubsection{Multi-step Training}
The aforementioned analysis does not involve the third training step and RSS. As shown in Table~\ref{tab:mAP}, although it only brings $0.14\%$ mAP improvement in the second training stage, the RSS makes it possible to conduct the third training step. Without the association loss, we obtain $65.13\%$ as an mAP after the final step, but $\mathcal L_{asso}$ can make further improvement.

\begin{figure}[!t]
\begin{center}
\includegraphics[width=7cm]{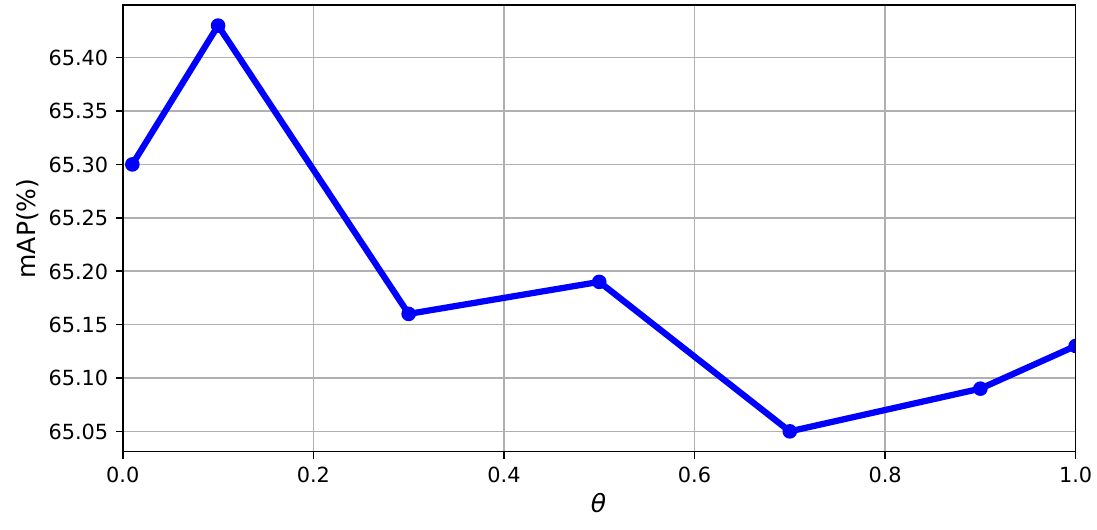}    
\caption{Detection performance vs. $\theta$. This figure shows model performance as a function of the confidence threshold $\theta$ in the association loss.}
\label{fig:asso}
\end{center}
\end{figure}

\begin{figure*}[!t]
\begin{center}
\includegraphics[width=16cm]{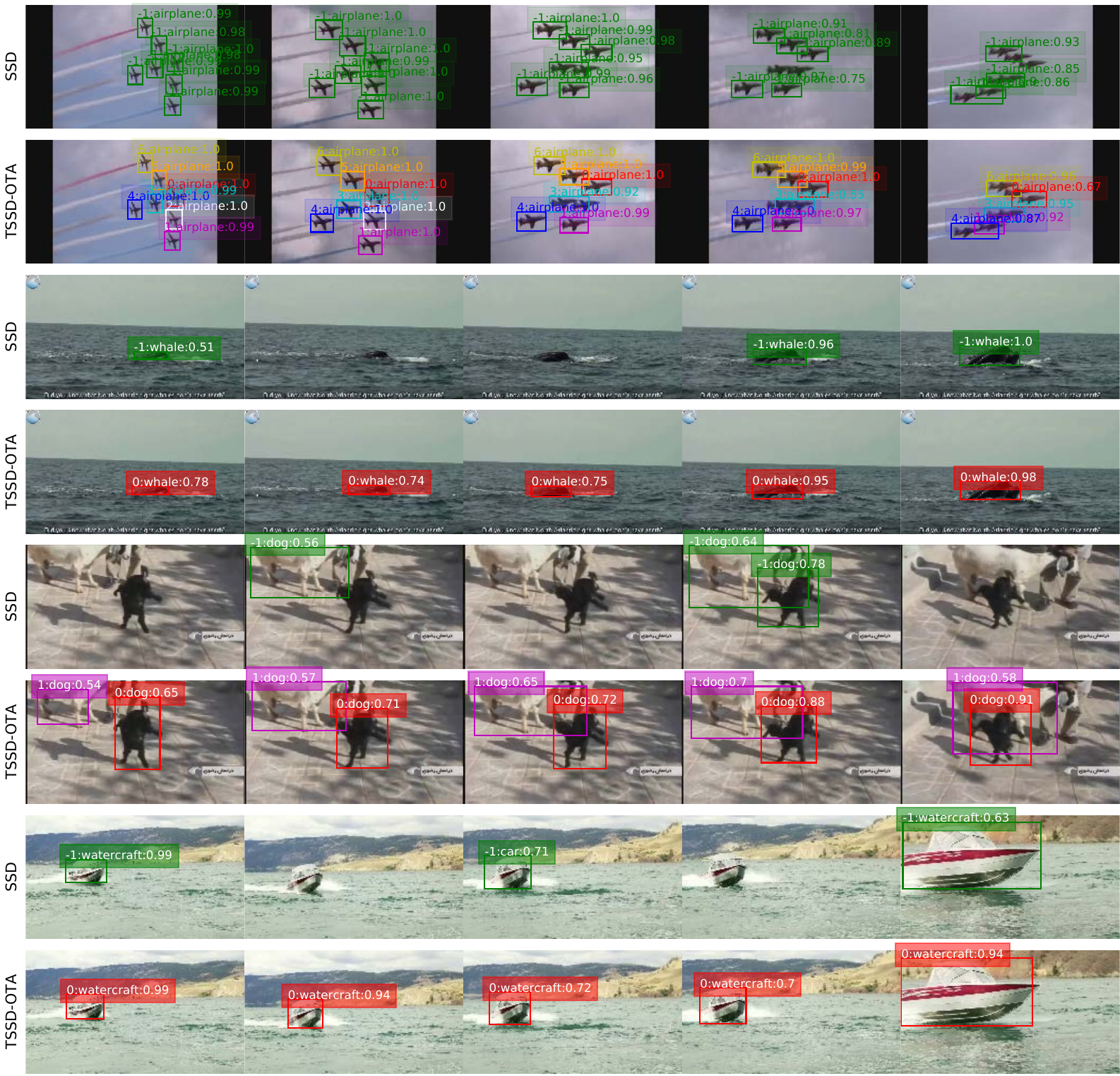}    
\caption{Demonstration results on the ImageNet VID validation dataset. The proposed TSSD(-OTA) can handle a variety of scenes with multiple objects more accurately. In addition, different from traditional detectors, our approach has the capability of identifying. The detection results are shown as ``\emph{ID:class:conf}''. \emph{ID}=-1 means the identity is not generated.}
\label{fig:comp}
\end{center}
\end{figure*}

The association loss is adopted in the third training step, where $sp=1$ assuring the training data is highly associated. In consideration of the employed NMS, there are three parameter settings for computing the association loss, i.e., confidence threshold $\theta$, IoU threshold, and top $k$ retained boxes. Because of the NMS, the number of retained boxes is relatively insensitive, and the IoU threshold could be consistent with that in the inference phase. Thus, the major implication of $\mathcal L_{asso}$ is $\theta$, which indicates what kind of objects should be involved. As depicted in Fig.~\ref{fig:asso}, the mAP generally decreases as $\theta$ increasing. That is, overmuch invalid boxes are considered when $\theta=0.01$, whereas the $\mathcal L_{asso}$ gradually loses efficacy as the decrease of involved objects. Hence, taking into account almost all positive samples, $\mathcal L_{asso}$ is the most effective when $\theta=0.1$. Consequently, the multi-step training brings $0.8\%$ mAP improvement, and the TSSD achieves $65.43\%$ in mAP.

\subsubsection{Comparison with Other Architectures}
We also compare the TSSD against several prior and contemporary approaches. As shown in Table~\ref{tab:comp}, their components and performances have been summarized. Most methods are based on a two-stage detector with RPN for region proposal, and few approaches successfully adopt attention or LSTM for temporal coherence. Additionally, tracking employed in TCNN \cite{bib:TCNN} and D\&T \cite{bib:DT} is a good idea for enhancing recall rate, but it affects time efficiency to some extent. In terms of the backbone, some methods based on the ResNet-101 can achieve higher mAP, but this deeper backbone is not suited to our task. Referring to \cite{bib:DSSD}, when it deals with small-size images in SSD, the ResNet-101 has an ignorable advantage in accuracy. The main reason is that two-stage approaches usually leverage large input size (e.g., $1000\times 600$), so deeper backbone can learn these richer visual features more effectively. However, our one-stage detector uses $300\times 300$ input image and relatively insufficient visual features for high inference speed, so the advantage of ResNet-101 is negligible. On the other hand, some approaches process video sequences in a batch mode (i.e., offline method), where future frames are also utilized for current frame detection. Hence, such non-causal methods are prohibitive for real-world applications. Because of its real-time and online characteristic, the TSSD is able to detect objects temporally for real-world applications. According to the authors' knowledge, our method has the following merits:
\begin{itemize}
  \item The TSSD is the first temporal one-stage detector achieving above 65\% mAP with such small input image and VGG-16 as a backbone.
  \item The TSSD-OTA is a unified framework, where a real-time online detector is capable of detecting and identifying objects. This framework is absent in the existing video detection approaches.
\end{itemize}

\subsubsection{Qualitative Results}
We show some qualitative results on the ImageNet VID validation set in Fig.~\ref{fig:comp}. We only display the detected bounding boxes with $>0.5$ confidence score. Different colors of the bounding boxes indicate different object identity. The proposed TSSD works better with precision and temporal stability.

\subsection{Tracking Performance on 2DMOT15 Dataset}
We employ the 2DMOT15 dataset to jointly investigate the TSSD-OTA, whose metrics are designed for tracking. The MOTA considers the comprehensive tracking performance, and the MOTP measures the tightness of the tracking results and ground truth. the FP, PN denote the total number of false positives and false negatives, respectively. For each trajectory, if more than $80\%$ of positions are successfully tracked, the MT increases by $1$. On the other hand, if more than $80\%$ of positions are lost, the ML increases by $1$. Finally, the IDS counts ID switch times.

\subsubsection{Parameter Analysis}
\begin{figure} \centering
\subfigure[] { \label{fig:match}
\includegraphics[width=7cm]{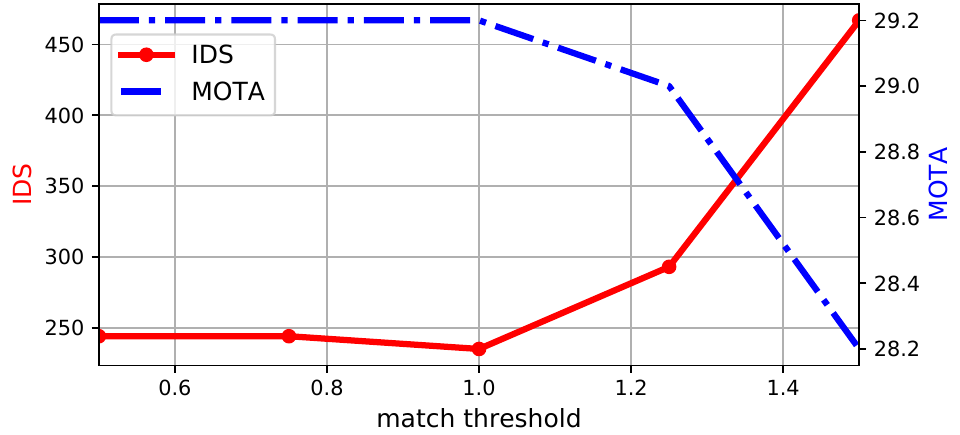}
}
\subfigure[] { \label{fig:len}
\includegraphics[width=7cm]{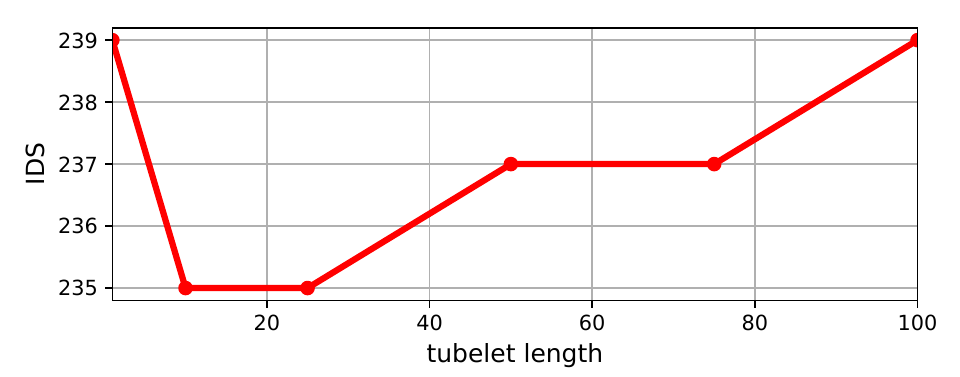}
}
\caption{Tracking performance vs. match threshold $\mathcal T$ and tubelet length $tub\_len$. }
\label{fig:ota_param}
\end{figure}

There are three parameters in the OTA, i.e., match threshold $\mathcal T$, tubelet generation score $\mathcal G$, and tubelet length $tub\_len$. To track each detected object, $\mathcal G$ should equal to detector's confidence threshold $conf$, and $\mathcal T,tub\_len$ will impact the tracking performance. We let $conf=\mathcal G=0.3, tub\_len=10$ to investigate the effect of $\mathcal T$. The change of MOTA and IDS is employed to represent the variation of tracking performance here. Referring to Fig.~\ref{fig:match}, the MOTA and IDS are strongly influenced by $\mathcal T$, and $\mathcal T=1.0$ is optimal as for our validation set. That is, if $\mathcal T$ is too high, the number of the failed match will increase. On the other hand, lower $\mathcal T$ causes the false match. Both failed and false match can weaken tracking performance. However, as shown in Fig.~\ref{fig:match}, failed match has a greater impact, because $\mathcal S^{max}_{obj}$ prevents the OTA from overmuch false match. as illustrated in Fig.~\ref{fig:len}, the IDS also first decreases and then increases as $tub\_len$ rising ($\mathcal T=1.0$). If $tub\_len=1$, the match process only relies on the most recent object, so it is unreliable for some emergencies (e.g., occlusion). However, an oversized $tub\_len$ would retain remote information, which may result in inaccurate.

We also unveil the effectiveness of attention maps for IDS with the parameters of $conf=\mathcal G=0.3, tub\_len=10, \mathcal T=1.0$. Note that the attention map for \emph{Conv11\_2}, whose size is $1\times 1$, is not qualified for identification, so it is not been involved in this test. Referring to Table~\ref{tab:ids}, the IDS is equal to $550$ when $\mathcal S$ is computed with IoU alone ($\mathcal T=0.5, \mathcal S=o$). The validity of attention maps is evident, and IDS drops about $60\%$ as a result. In addition, the effectiveness gradually declines as visual scale increasing, since high-level features usually contain more within-class-similar information. Further, we also combine multi-scale attention maps to compute $\mathcal S$, and the best result is obtained when attention maps in low-level AC-LSTM are employed.

\begin{table}[!t]
\renewcommand{\arraystretch}{1}
\caption{Effectiveness of attention maps for IDS. The checkmark indicates that corresponding attention map is employed for $\mathcal S$.}
\label{tab:ids}
\centering
\begin{tabular}{c c c c c | c}
\Xhline{1.5pt}
\emph{Conv4\_3}  & \emph{Conv7}    &  \emph{Conv8\_2} & \emph{Conv9\_2} & \emph{Conv10\_2}  & IDS  \\
\hline
           &            &            &            &            & $550$  \\
\checkmark &            &            &            &            & $236$  \\
           & \checkmark &            &            &            & $240$  \\
           &            & \checkmark &            &            & $238$  \\
           &            &            & \checkmark &            & $244$  \\
           &            &            &            & \checkmark & $251$  \\
\checkmark & \checkmark &            &            &            & $236$  \\
\checkmark & \checkmark &\checkmark  &            &            & $235$  \\
\checkmark & \checkmark &\checkmark  &\checkmark   &           & $236$  \\
\checkmark & \checkmark &\checkmark  &\checkmark  &\checkmark  & $237$  \\
\Xhline{1.5pt}
\end{tabular}
\end{table}

\begin{table*}[!t]
\renewcommand{\arraystretch}{1.2}
\caption{Tracking performance on the 2DMOT15 validation set.}
\label{tab:MOT}
\centering
\begin{tabular}{c | c c c c c c c c c}
\Xhline{1.5pt}
Method & MOTA $\uparrow$ & MOTP $\uparrow$ & MT $\uparrow$ & ML $\downarrow$ & FP $\downarrow$ & FN $\downarrow$ & IDS $\downarrow$  & FPS $\uparrow$  \\
\hline
ACF-MDP \cite{bib:ACF,bib:MDP}     & 26.7          & 73.6          & 12.0\%      & 51.7\%        & 3290     & 13491     & 133  & $\sim<1/1$ \\
SSD-ALSTM \cite{bib:ALSTM}         & 38.6          & 74.2          & 14.9\%      & 46.8\%        & 788      & 13253     & 154  & $\sim9/12$ \\
FasterRCNN-SORT \cite{bib:SORT}    & 34.0          & 73.3          & 20.5\%      & 32.1\%        & 3311     & 11660     & 274      & $\sim5/220$ \\
\hline
TSSD-SORT                          & 25.6          & 72.4          & 6.4\%    & 48.9\%   & 798     & 16121   & 260      & $\sim 28/420$ \\
TSSD-OTA(propsed)                  & 29.2          & 71.9          & 12.4\%   & 34.6\%   & 1482    & 14624   & 235     & $\sim27/270$ \\
\Xhline{1.5pt}
\end{tabular}
\end{table*}

\begin{figure*}[!t]
\centering
\includegraphics[width=16cm]{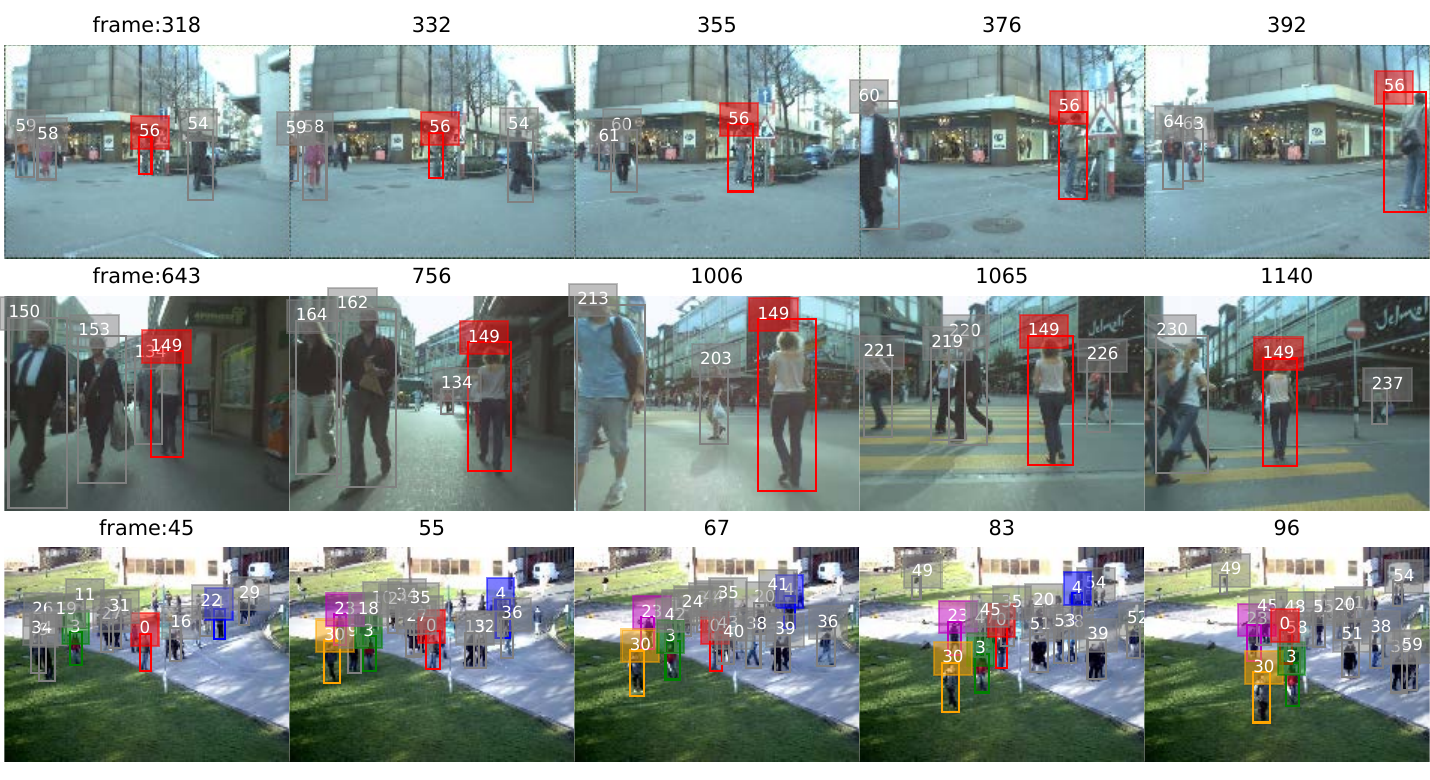}    
\caption{Demonstration results on the 2DMOT15 validation dataset. The TSSD-OTA can track pedestrians in a variety of multi-target scenes. The identities for each object are denoted on boxes, and some representative results are demonstrated in red, green, blue, or orange.}
\label{fig:ota}
\end{figure*}

\subsubsection{Tracking Results}
Under the conditions of $tub\_len=10,\mathcal T=1.0, conf=\mathcal G=0.3$, we use the TSSD-OTA to track pedestrians in 2DMOT15 dataset. Referring to Table~\ref{tab:MOT}, the FPS is described in the form of \emph{overall speed / tracking component speed}. A conclusion can be drawn based on the comparison with prior and contemporary approaches that the proposed TSSD-OTA can run at frame-rates beyond real time while maintaining comparable tracking performance. Concerning the MT and ML, the FasterRCNN-SORT performs better because the two-stage detector works well on recall rate. In case of MOTA, the SSD-ALSTM is better owing to their LSTM-based tracking component. However, their solutions come with a major drawback, i.e., high time cost. There are two main reasons causing the proposed TSSD-OTA cannot be equal to the compared methods on some criteria. i) There are vast small objects in the 2DMOT15 dataset, but our detector with such small input image is not adept at dealing them; ii) Unlike most methods, our model is optimized without any objective related to identity or association. The proposed association loss is also not adopted in this experiment, and we will explain the reason in Section~\ref{sec:Dis}.

Then, employing the TSSD as the detector, we compare the OTA and SORT \cite{bib:SORT} comprehensively. As shown in Table~\ref{tab:MOT}, we achieve better indexes on MOTA, MT, ML, FN, and IDS. That is, the OTA generates better tracking performance while keeping quite high processing speed.

\subsubsection{Qualitative Results}
As schematically illustrated in Fig.~\ref{fig:ota}, the proposed TSSD-OTA is able to track pedestrians in a variety of scenarios, and we demonstrate some typical results. In the first line of Fig.~\ref{fig:ota}, \#56 is in a small size at the beginning, then it becomes larger with the change of visual perspective. As a result, the TSSD-OTA can adapt to this continuous scale change. \#149 in the second line shuttles in the crowd, and it undergoes illumination changes. Still, this challenging target is tracked well by the TSSD-OTA. There are chaotic small objects in the third line of Fig.~\ref{fig:ota}, and they occlude each other. Nevertheless, our method works well in terms of \#0, \#3, \#4, \#23, \#30, and etc. Unfortunately, there are some failed cases appear in this scenario, e.g., ID switches (\#15$\to$\#39, \#26$\to$\#23, \#34$\to$\#30), false negatives (\#4,\#49).

\subsection{Discussion}
\label{sec:Dis}
We employ ImageNet VID and 2DMOT15 datasets for experiments, which are significantly different as for video amount (4000 v.s. 8), class number (30 v.s. 1), and metrics (mAP v.s. MOTA and etc.). In addition, 2DMOT15 contains vast small objects while VID covers multifarious scenarios. Hence, we discuss the following topics in detail.

\subsubsection{RSS}
As shown in Table~\ref{tab:mAP}, there are only $0.14\%$ mAP increase after RSS is adopted. However, the reduced data amount highlights the need for it. For example, the TSSD cannot be well trained without RSS on the 2DMOT15 dataset, resulting in that the MOTA drops $1.4$ after the third training step. Obviously, the phenomenon is caused by the diversity of training data. That is, due to a variety of scenarios in VID dataset, the data diversity can be preserved without the RSS. Unfortunately, this trick cannot be neglected when training 2DMOT15 dataset, since there are only $8$ employed videos. Since our association loss requires continuous sampling, it is not adopted during 2DMOT15 training.

\subsubsection{Association Loss}
It is not the first time that an association loss is designed. For instance, Lu \emph{et al.} also exploited one to train an LSTM \cite{bib:ALSTM}. As opposed to previous design, our association loss focuses on global classification rather than each associated object pair, forgoing the need to introduce identity ground truth labels to the training process. It is necessary because detection datasets usually do not involve the identity label, or collecting ID-labelled data is a costly work. Although the OTA can generate objects' identities, it is not suitable for computing association loss. That is, the OTA also cannot be supervised for lack of ID labels, so its errors are directly uncorrectable during training. Furthermore, the OTA is closely dependent on TSSD's results, and it cannot perform well if the TSSD is not well trained. If association loss is computed with the OTA, some match errors could occur, making the training process unstable.

\subsubsection{Trade-off Between Accuracy and Speed}
In our experiments, the average inference time of SSD, SSD+ConvLSTM, SSD+AC-LSTM are $0.022$~s, $0.026$~s, and $0.037$~s, respectively. Thus, the attention module has a bigger impact on the detection speed because it conducts a $3$-layer convolution (ConvLSTM only needs a $1$-layer convolution). Thus, simplifying the attention module could be in favor of inference speed, but obviously, the attention performance could also decline. Besides, special designs for ConvLSTM can increase the speed as well, e.g., Bottleneck-LSTM \cite{bib:LSTM_SSD}. The method for integrating temporal information is an essential step towards high detection accuracy. On one hand, single-object tracking \cite{bib:DT} is more effective than LSTM. (Note that multi-object tracking or tracking-by-detection methods, like OTA, cannot be leveraged for this purpose because it needs detection results at every time steps.) However, single-object tracking methods usually have adverse effects on inference speed. On the other hand, it is well known that SSD's performance comes with a series of data augmentation, e.g., SSD randomly expands and crops the original training images with additional random photometric distortion \cite{bib:SSD}. However, most existing data augmentation methods do not suitable for sequential learning (i.e., they would change the spatiotemporal relation), so the second and third training steps do not include any data augmentation. Thus, exploiting a sequential data augmentation during training could further improve the detection accuracy. By and large, the TSSD achieves a reasonable trade-off between accuracy and speed.

\subsubsection{OTA}
As for the OTA, there have been some similar methods, e.g., \cite{bib:OnlineACT,bib:OfflineACT} analyzed spatiotemporal action based on tubelets. These approaches and the OTA have a similar purpose, i.e., linking detected boxes to existing tubelets for data association. However, we have two key advantages: 1) They first select ``candidates'' using IoU threshold. On the contrary, we treat the IoU threshold as a weight that impacts the similarity score (see (\ref{eqn:simi})), so the OTA could be more computationally efficient; 2) Match score in existing methods is usually based on the mean of the confidence scores of the tube's member detection boxes. These confidence scores could suffer from unstable fluctuations because pixel-level changes could significantly impact the confidence score. Besides, the method in \cite{bib:OnlineACT,bib:OfflineACT} only deals with a small number of objects, but dozens of objects with similar confidence scores should be considered for multi-object tracking task. Thus, the method proposed by \cite{bib:OnlineACT,bib:OfflineACT} is not discriminative enough for tracking numerous objects. Conversely, we use attention vector from a temporal detector to describe objects' saliency mass, and the match between boxes and tubelets is based on the similarity of the attention vector. Thus, the OTA is in favor of feature-level match.

\subsection{A Real-World Application}
\begin{figure}[!t]
\centering
\includegraphics[width=8cm]{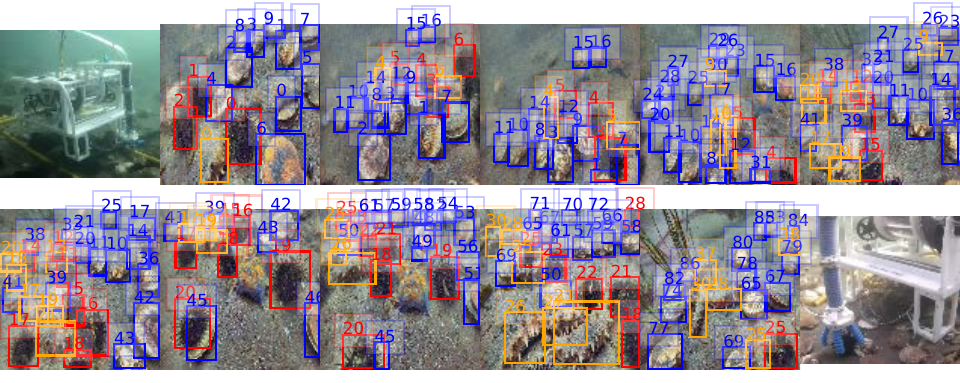}
\caption{Undersea object grasping and amount statistics. Sea-urchins, scallops, and sea-cucumbers are demonstrated in red, blue and orange. Objects' identities are denoted on the top of boxes. Besides counting the number of objects, the TSSD-OTA can locate each target individual for grasping.}
\label{fig:uw}
\end{figure}

As shown in Fig.~\ref{fig:uw}, we conducted practical experiments on the seabed using a remotely operated vehicle (ROV). Equipped with a camera as visual guidance, the ROV is $0.68$ m in length, $0.57$ m in width, $0.39$ m in height, and $50$ kg in weight. The test venue is located in Zhangzidao, Dalian, China, where the water depth is approximately $10$ m. For humans, managing seafloor products (e.g., sea-urchins, scallops, sea-cucumbers, and etc.) is very difficult. Moreover, their collection has been a big problem. To that end, we utilize the ROV for this laborious and dangerous job instead of humans. Further, the TSSD-OTA is of crucial importance in this maneuver. On one hand, our proposed method can accurately count the amount for each category of sea-products when the ROV travels straightly. On the other hand, distinguished from traditional detection approaches, the TSSD-OTA is capable of locating each target individual for grasping. The full video demonstration is available at \url{https://youtu.be/v3LG-O-abWI}.

\section{Conclusion and Future Work}
\label{sec:CON}
This paper has aimed at temporally detecting and identifying objects in real time for real-world applications. A creative TSSD is proposed. Differing from existing video detection methods, the TSSD is a temporal one-stage detector, and it can perform well in terms of both detection precision and inference speed. To efficiently integrate pyramidal feature hierarchy, an LH-TU is proposed, in which the high-level features and low-level features share their respective temporal units. Furthermore, we design an AC-LSTM as a temporal analysis unit, where the temporal attention mechanism is responsible for background suppression and scale suppression. A novel association loss function and multi-step training are also designed for sequence learning. In addition, the OTA algorithm equips the TSSD with the ability of identification with low computational costs. As a result, the TSSD-OTA sees considerably enhanced detection precision, tracking performance, and inference speed. Finally, our proposed approaches have been employed for real-world applications.

In the future, we plan to further improve the inference speed of the TSSD-OTA. Besides, the TSSD-OTA will be used for robotic visual navigation under dynamic environments.

\ifCLASSOPTIONcaptionsoff
  \newpage
\fi

\end{document}